# Counting Markov Blanket Structures

Shyam Visweswaran, MD, PhD and Gregory F. Cooper, MD, PhD

Department of Biomedical Informatics and the Intelligent Systems Program,
University of Pittsburgh School of Medicine, Pittsburgh, PA

## Abstract

Learning Markov blanket (MB) structures has proven useful in performing feature selection, learning Bayesian networks (BNs), and discovering causal relationships. We present a formula for efficiently determining the number of MB structures given a target variable and a set of other variables. As expected, the number of MB structures grows exponentially. However, we show quantitatively that there are many fewer MB structures that contain the target variable than there are BN structures that contain it. In particular, the ratio of BN structures to MB structures appears to increase exponentially in the number of variables.

## 1. INTRODUCTION

An important task in machine learning and data mining is to characterize how the target variable is influenced by other variables in the domain. For example, the goal of a classification or regression algorithm is to learn how the target variable can be best predicted from the other variables. One approach to characterizing the influence on a target variable is to identify a subset of the variables that shields that variable from the influence of other variables in the domain. Such a subset of variables is called a *Markov blanket* (MB) of the target variable. As introduced by Pearl (Pearl 1988), the term MB refers to any subset of variables that shield the target variable from the influence of other variables in the domain. The *minimal Markov blanket* or the *Markov boundary* is a minimal subset of variables that shield the target variable from the influence of other variables in the domain (Pearl 1988). Many authors refer to the minimal Markov blanket as the Markov blanket, and we follow this convention.

The notion of MBs and the identification of MBs have several uses in machine learning and data mining. The identification of the MB of a variable is useful in the feature (variable) subset selection problem (Koller and Sahami 1996; Ioannis, Constantin et al. 2003). Feature selection is useful in removing irrelevant and redundant variables and in improving the accuracy of classifiers (Dash and Liu 1997). MB identification can be used for guiding the learning of the graphical structure of a Bayesian network (BN) (Margaritis and Thrun 1999). Identifying MBs is also useful in the discovery of causal structures where the goal is to discover direct causes, direct effects, and common causes of a variable of interest (Mani and Cooper 2004).

Several methods of learning BNs from data have been described in the literature (Cooper and Herskovits 1992; Heckerman, Geiger et al. 1995). While the MB of a target variable can be extracted from the BN learned by such methods, methods that learn only the MB of a target variable may be more efficient. The general motivation for searching over the space of MBs is that it is smaller than the space of BNs. Given a set of domain variables, we quantify the size of the MB search space and compare it to the size of the BN search space. We derive a formula for efficiently determining the number of MB structures of a target variable given other variables, and we also compute the ratio of BN structures to MB structures for a given number of variables.

The remainder of the paper is organized as follows. We review briefly BNs and MBs in Sections 2 and 3. In Section 4, we derive a formula for counting MBs. In Section 5 we compare the number of BN structures and MB structures and in Section 6 we provide a summary of the results.

## 2. BAYESIAN NETWORKS

A Bayesian network is a probabilistic model that combines a graphical representation (the BN structure) with quantitative information (the BN parameterization) to represent a joint probability distribution over a set of random variables. More specifically, a BN representing the set of random variables $X_i \,\epsilon\, \mathbf{X}$ in some domain consists of a pair $(G, \mathbf{\Theta}_G)$. The first component $G$ that represents the BN structure is a directed acyclic graph (DAG) that contains a node for every variable in $\mathbf{X}$ and an arc between a pair of nodes if the corresponding variables are directly probabilistically dependent. Conversely, the absence of an arc between a pair of nodes denotes probabilistic independence between the corresponding variables. Each variable $X_i$ is represented by a corresponding node labeled $X_i$ in the BN graph, and in the this paper we use the terms node and variable interchangeably. The second component $\Theta_G$ represents the parameterization of the probability distribution over the space of possible instantiations of *X* and is a set of local probabilistic models that encode quantitatively the nature of dependence of each node on its parents. For each node $X_i$ there is a local probability distribution (that may be discrete or continuous) defined on that node for each state of its parents. The set of all the local probability distributions associated with all the nodes comprises a complete parameterization of the BN. Note that the phrase BN structure refers only to the graphical structure *G*, while the term BN (model) refers to both the structure *G* and a corresponding set of parameters $\Theta_G$.

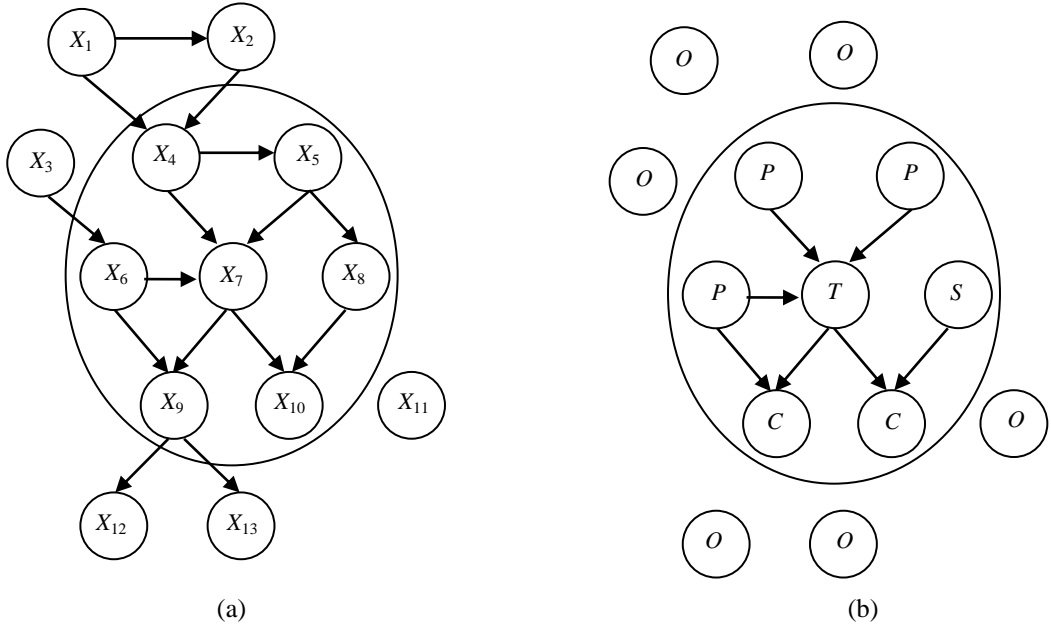

**Figure 1:** (a) A BN structure for a domain with 13 variables. The MB of $X_7$ includes nodes within the circled region, namely, the parents, children and the spouses of $X_7$. (b) An example of a MB structure demonstrating various node types. *T* is the *target* node, *P* is a *parent* node, *C* is a *child* node, *S* is a *spousal* node, and *O* is an *other* node. Note that *other* nodes are not part of the MB structure and no arcs are allowed between parents or between a parent and a spouse (see text).

The terminology of kinship is used to denote various relationships among nodes in a BN. These kinship relations are defined along the direction of the arcs. Predecessors of a node $X_i$ in *G*, both immediate and remote, are called the ancestors of $X_i$. In particular, the immediate predecessors of $X_i$ are called the parents of $X_i$. In a similar fashion, successors of $X_i$ in *G*, both immediate and remote, are called the descendants of $X_i$, and, in particular, the immediate successors are called the children of $X_i$. A node $X_j$ is termed a spouse of $X_i$ if $X_j$ is a parent of a child of $X_i$. The set of nodes consisting of a node $X_i$ and its parents is called the family of $X_i$. An example of a BN structure is given in Figure 1(a).

A formula for efficiently determining the number of BN structures was first developed by Robinson (Robinson 1971; Robinson 1976), and independently by Stanley (Stanley 1973). The number of BN structures that can be constructed from *n* variables is given by the following recurrence formula, where *BN(n)* is the number of DAGs that can be constructed given *n* nodes:

$$BN(n) = \sum_{k=1}^{n}(-1)^{k+1}\binom{n}{k}2^{k(n-k)}BN(n-k), n > 0$$

$$BN(0) = 1 \qquad (1)$$

$$where \binom{n}{k} = \frac{n!}{k!\,(n-k)!}$$

and denotes the number of ways to choose *k* nodes from *n* nodes.

Equation 1 is computed using dynamic programming, namely, the values of *BN(\*)* are computed only once, then cached for whenever they are needed again.

The time complexity of computing Equation 1 is $O(n^2)$ where *n* is the number of variables. From Equation 1, it can be seen that *BN(n)* is computed from four terms: the first term takes $O(1)$ time, the second term which gives the number of ways to choose *k* nodes from *n* nodes takes $O(n^2)$ time, the third term takes $O(n^2)$ time and the fourth term is retrieved in $O(1)$ time from the cache. This analysis assumes that all operations can be done in $O(1)$ time, which is probably a reasonable assumption for $n$ small enough that the operations do not cause an arithmetic overflow or underflow. As $n$ gets large, special routines may be needed to handle operations on $n$, which will increase the time complexity.

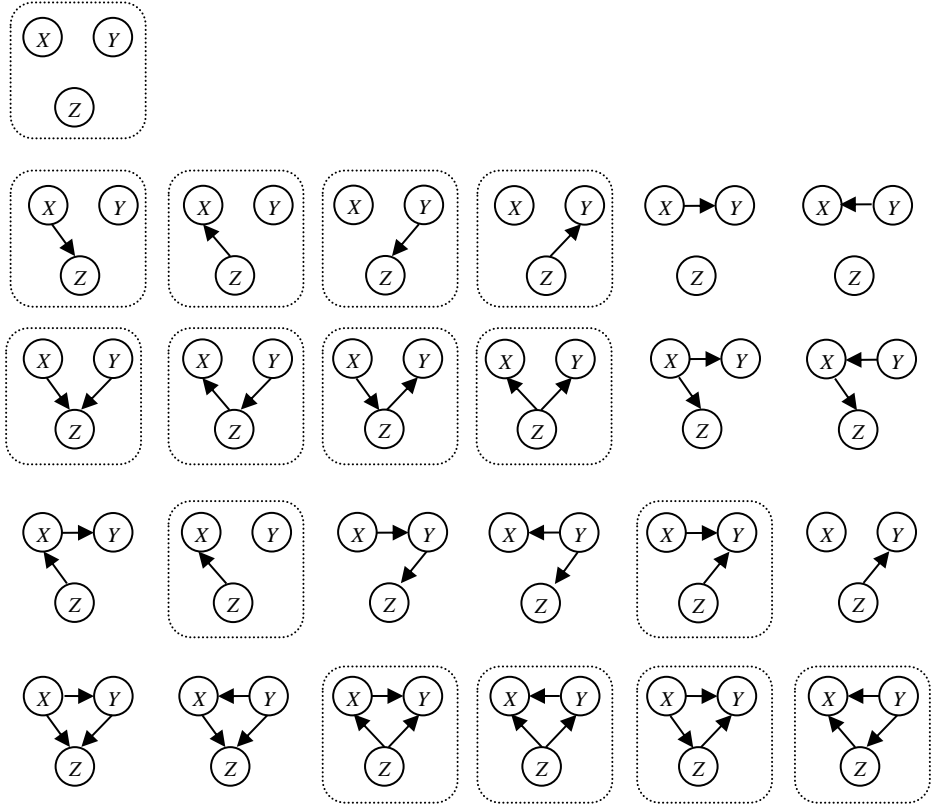

**Figure 2:** The 25 BN structures for a domain containing three variables. The 15 MB structures with respect to variable Z are indicated by dashed boxes around them.

## 3. MARKOV BLANKETS

The MB of a variable $X_i$, denoted by $MB(X_i)$, defines a set of variables such that conditioned on $MB(X_i)$, $X_i$ is conditionally independent of all variables outside of $MB(X_i)$. The minimal MB is a minimal set $MB(X_i)$ conditioned on which $X_i$ is conditionally independent of all variables outside of $MB(X_i)$. As mentioned in the introduction, we shall refer to the minimal MB as the MB. Analogous to BNs, a MB structure refers only to the graphical structure while a MB refers to both the structure and a corresponding set of parameters. The MB structure consists of the parents, children, and children's parents of $X_i$, as illustrated in Figure 1(b). Different MBs that entail the same factorization for the conditional distribution $P(X_i | MB(X_i))$ belong to the same *Markov equivalence class*. We define a *MB structure* specifically as a representative member of a Markov equivalence class that has no arcs between parents or between a parent and a spouse.

With respect to a MB structure, the nodes of the BN can be categorized into five groups: (1) the target node, (2) parent nodes of the target, (3) child nodes of the target, (4) spousal nodes, which are parent nodes of the children, and (5) other nodes, which are not part of the MB. The node under consideration is called the *target* node. A *parent* node is one that has an outgoing arc to the target node and may have additional outgoing arcs to one or more child nodes. A *child* node is one that has an incoming arc from the target node, may have additional incoming arcs from parent nodes, spousal nodes and other child nodes, and may have outgoing arcs to other child nodes. A *spousal* node is one that has outgoing arcs to one or more child nodes and has neither an incoming arc from the target node nor an outgoing arc to it. An *other* node is one that is not in the MB and is considered to be a potential spousal node, as we explain below. An example demonstrating the various types of nodes in a MB structure is given in Figure 1(b). As mentioned earlier, in a MB structure we disallow arcs between parent nodes, arcs between parent nodes and spousal nodes and arcs between child nodes and other nodes.

# 4. COUNTING MARKOV BLANKET STRUCTURES

We now derive of a formula for counting the number of distinct MB structures with respect to a target variable. The number of possible MB structures for a domain with $n$ variables is given by the following equation:

$$MB(n) = \sum_{n_p=0}^{n-1} \sum_{n_{so}=0}^{n-n_p-0} \left( \frac{(n-1)!}{n_p! \, n_c! \, n_{so}!} \right) 2^{n_c n_p} 2^{n_c n_{so}} BN(n_c), n > 1$$

$$MB(1) = 1$$

(2)

where $n_c = n - n_p - n_{so} - 1$

where, $n_p$ is number of parent nodes, $n_c$ is the number of child nodes, $n_{so}$ is the number of spousal and other nodes, and $BN(n_c)$ is the number of DAGs that can be constructed with $n_c$ nodes and is given by Equation 1.

Equation 2 is derived as follows. The terms inside the double summation count the number of MB structures for a specified number of $n_p$, $n_c$ and $n_{so}$ nodes. The first term gives the number of ways $n - 1$ can be partitioned into $n_p$ parent nodes, $n_c$ child nodes and $n_{so}$ spousal and other nodes; this term is sometimes called the *multiplicity*. The second term gives the number of distinct MB structures that differ only in the presence or absence of arcs from parent nodes to child nodes. Each parent node can have an arc to none, one or more child nodes for a total of $2^{n_c}$ distinct MB structures. For $n_p$ parent nodes, the number of distinct MB structures that differ only in the presence or absence of arcs from parent nodes to child nodes is $2^{n_c n_p}$. The third term gives the number of distinct MB structures that differ only in the presence or absence of arcs from spousal and other nodes to child nodes. This derivation is similar to the derivation of the previous term. Each spousal or other node can have an arc to none, one, or more child nodes for a total of $2^{n_{so}}$ distinct MB structures. For $n_{so}$ spousal or other nodes, the number of distinct MB structures that differ only in the presence or absence of arcs from spousal or other nodes to child nodes is $2^{n_c n_{so}}$. The fourth and last term gives the number of DAGs that can be constructed with $n_c$ child nodes. The summation is carried over all possible values of $n_p$ and $n_{so}$; selection of particular values for $n_p$ and $n_{so}$ determines the value of $n_c$ and hence no explicit summation is required over the values of $n_c$. Note that the number of parent nodes, the number of child nodes, the number of spousal and other nodes and the single target node together add up to $n$ nodes. Excluding the target node, the sum of the nodes is $n-1$.

**Table 1:** Number of BN structures $BN(n)$ and MB structures $MB(n)$ as a function of number of nodes $n$. The number of MB structures is with respect to a single target node and is not a count of all MB structures for all nodes. The last column gives the ratio of the two types of structures. Both $BN(n)$ and $MB(n)$ are exponential in $n$.

| $n$ | $BN(n)$ | $MB(n)$ | Ratio $BN(n) / MB(n)$ |
|---|---|---|---|
| 1 | 1 | 1 | 1.0 |
| 2 | 3 | 3 | 1.0 |
| 3 | 25 | 15 | 1.66666666667 |
| 4 | 543 | 153 | 3.54901960784 |
| 5 | 29,281 | 3,567 | 8.20885898514 |
| 6 | 3,781,503 | 196,833 | 19.2117327887 |
| 7 | 1,138,779,265 | 25,604,415 | 44.4758946846 |
| 8 | 783,702,329,343 | 7,727,833,473 | 101.412942202 |
| 9 | 1,213,442,454,842,881 | 5,321,887,813,887 | 228.009777222 |
| 10 | 4,175,098,976,430,598,143 | 8,241,841,773,665,793 | 506.573541580 |
| 11 | 31,603,459,396,418,917,607,425 | 28,359,559,029,362,676,735 | 1,114.38472522 |
| 12 | 521,939,651,343,829,405,020,504,063 | 214,672,167,825,864,945,784,833 | 2,431.33358474 |
| 13 | 1.867660E+031 | 3.545390E+027 | 5,267.85556534 |
| 14 | 1.439428E+036 | 1.268651E+032 | 11,346.1282090 |
| 15 | 2.377253E+041 | 9.777655E+036 | 24,313.1173477 |
| 16 | 8.375667E+046 | 1.614805E+042 | 51,867.9742260 |
| 17 | 6.270792E+052 | 5.689370E+047 | 110,219.439109 |
| 18 | 9.942120E+058 | 4.259584E+053 | 233,405.867102 |
| 19 | 3.327719E+065 | 6.753420E+059 | 492,745.716894 |
| 20 | 2.344880E+072 | 2.260432E+066 | 1,037,359.40236 |
| 21 | 3.469877E+079 | 1.592816E+073 | 2,178,454.74390 |
| 22 | 1.075823E+087 | 2.356996E+080 | 4,564,381.36751 |

As an example, for a domain containing three variables, the number of BN structures is 25 and the number of MB structures is 15 as given by Equations 1 and 2 respectively. Figure 2 shows all 25 BN structures and indicates the 15 MB structures with respect to one of the domain variables.

The time complexity of computing Equation 2 is $O(n^2)$ where $n$ is the number of domain variables. From Equation 2, it can be seen that $MB(n)$ is computed from four terms: the first term takes $O(n^2)$ time, the second and third terms take $O(n)$ time each and the fourth term takes $O(n^2)$ time as derived previously for Equation 1.

Of note the number of distinct possible MB structures is the same for any variable in a given domain, and thus, Equation 2 applies to an arbitrary domain variable that is considered the target variable.

## 5. A COMPARISON OF THE NUMBER OF BAYESIAN NETWORK AND MARKOV BLANKET STRUCTURES

Table 1 gives the values of $BN(n)$ and $MB(n)$ for $n$ ranging from 0 to 22 where $BN(n)$ and $MB(n)$ are computed from Equations 1 and 2 respectively. It can be seen from the table that while there are fewer MB structures of the target variable than there are BN structures, the number of MB structures is exponential in the number of variables. Thus, as is generally appreciated, exhaustive search in the space of MB structures will usually be infeasible for domains containing more than a few variables and heuristic search is appropriate.

The last column in Table 1 gives the values of the ratio of $BN(n)$ to $MB(n)$ for $n$ ranging from 1 to 22. It appears that this ratio is increasing exponentially in $n$. Thus, searching in the space of MB structures is likely to be relatively much more efficient than searching in the space of BN structures.

## 6. SUMMARY

We have presented a formula for efficiently determining the number of MB structures without enumerating them explicitly. Although there are fewer MB structures than there are BN structures, the number of MB structures is exponential in the number of variables. However, the ratio of BN structures to MB structures appears to increase exponentially in the number of domain variables. Thus, while searching exhaustively in the space of MB structures will usually be infeasible, searching in the space of MB structures is likely to be more efficient than searching in the space of BN structures. Thus, for algorithms that need only learn the MB structure this paper quantifies the degree to which it is preferable to search in the space of MB structures rather than in the space of BN structures.

## References

Cooper, G. F. and E. Herskovits (1992). A Bayesian method for the induction of probabilistic networks from data. Machine Learning **9**(4): 309-347.

Dash, M. and H. Liu (1997). Feature selection for classification. Intelligent Data Analysis **1**: 131–156.

Heckerman, D., D. Geiger, et al. (1995). Learning Bayesian networks - the combination of knowledge and statistical data. Machine Learning **20**(3): 197-243.

Ioannis, T., F. A. Constantin, et al. (2003). Time and sample efficient discovery of Markov blankets and direct causal relations. Proceedings of the Ninth ACM SIGKDD International Conference on Knowledge Discovery and Data Mining, Washington, D.C., ACM.

Koller, D. and M. Sahami (1996). Toward optimal feature selection. Proceedings of the Thirteenth International Conference on Machine Learning.

Mani, S. and G. F. Cooper (2004). A Bayesian local causal discovery algorithm. Proceedings of the World Congress on Medical Informatics, San Francisco, CA.

Margaritis, D. and S. Thrun (1999). Bayesian network induction via local neighborhoods. Proceedings of the 1999 Conference on Advances in Neural Information Processing Systems, Denver, CO, MIT Press.

Pearl, J. (1988). Probabilistic Reasoning in Intelligent Systems. San Mateo, California, Morgan Kaufmann.

Robinson, R. W. (1971). Counting labeled acyclic digraphs. New Directions in Graph Theory, Third Ann Arbor Conference on Graph Theory, University of Michigan, Academic Press.

Robinson, R. W. (1976). Counting unlabeled acyclic digraphs. Proceedings of the Fifth Australian Conference on Combinatorial Mathematics, Melbourne, Australia.

Stanley, R. P. (1973). Acyclic orientations of graphs. Discrete Mathematics **5**: 171-178.